\documentclass[fleqn,10pt]{wlscirep}
\usepackage[utf8]{inputenc}
\usepackage[T1]{fontenc}
\usepackage[svgnames,dvipsnames]{xcolor}
\usepackage{xspace}
\usepackage{array}
\usepackage{tabularx}
\title{A Framework for Longitudinal Health AI Agents}

\author[1,*]{Georgianna "Blue" Lin}
\author[2]{Rencong Jiang}
\author[1]{Noémie Elhadad}
\author[1]{Xuhai "Orson" Xu}
\affil[1]{Columbia University, Biomedical Informatics, New York, New York USA}
\affil[2]{Columbia University, Computer Science, New York, New York USA}

\affil[*]{gl2981@cumc.columbia.edu}

\newcommand{\coherence}{\textbf{\textcolor{Violet}{coherence}}\xspace}
\newcommand{\coherenceHistory}{\textit{\textcolor{Violet}{history}}\xspace}
\newcommand{\coherenceOrganization}{\textit{\textcolor{Violet}{organization}}\xspace}
\newcommand{\coherenceRelationship}{\textit{\textcolor{Violet}{relationship}}\xspace}
\newcommand{\coherencePersistence}{\textit{\textcolor{Violet}{persistence}}\xspace}
\newcommand{\continuity}{\textbf{\textcolor{Thistle}{continuity}}\xspace}
\newcommand{\continuityFollowup}{\textit{\textcolor{Thistle}{follow-up}}\xspace}
\newcommand{\continuityAlignment}{\textit{\textcolor{Thistle}{alignment}}\xspace}
\newcommand{\continuityAccountability}{\textit{\textcolor{Thistle}{accountability}}\xspace}
\newcommand{\adaptation}{\textbf{\textcolor{DarkSeaGreen}{adaptation}}\xspace}
\newcommand{\adaptationResponsiveness}{\textit{\textcolor{DarkSeaGreen}{responsiveness}}\xspace}
\newcommand{\adaptationPersonalization}{\textit{\textcolor{DarkSeaGreen}{personalization}}\xspace}
\newcommand{\adaptationReflexivity}{\textit{\textcolor{DarkSeaGreen}{reflexivity}}\xspace}
\newcommand{\agency}{\textbf{\textcolor{Peach}{agency}}\xspace}
\newcommand{\agencyNegotiation}{\textit{\textcolor{Peach}{negotiation}}\xspace}
\newcommand{\agencyTransparency}{\textit{\textcolor{Peach}{transparency}}\xspace}
\newcommand{\agencyEmancipation}{\textit{\textcolor{Peach}{emancipation}}\xspace}
\newcommand{\agencyProactivity}{\textit{\textcolor{Peach}{proactivity}}\xspace}

\newcommand{\Coherence}{\textbf{\textcolor{Violet}{Coherence}}\xspace}
\newcommand{\CoherenceHistory}{\textit{\textcolor{Violet}{History}}\xspace}
\newcommand{\CoherenceOrganization}{\textit{\textcolor{Violet}{Organization}}\xspace}
\newcommand{\CoherenceRelationship}{\textit{\textcolor{Violet}{Relationship}}\xspace}
\newcommand{\CoherencePersistence}{\textit{\textcolor{Violet}{Persistence}}\xspace}
\newcommand{\Continuity}{\textbf{\textcolor{Thistle}{Continuity}}\xspace}
\newcommand{\ContinuityFollowup}{\textit{\textcolor{Thistle}{Follow-up}}\xspace}
\newcommand{\ContinuityAlignment}{\textit{\textcolor{Thistle}{Alignment}}\xspace}
\newcommand{\ContinuityAccountability}{\textit{\textcolor{Thistle}{Accountability}}\xspace}
\newcommand{\Adaptation}{\textbf{\textcolor{DarkSeaGreen}{Adaptation}}\xspace}
\newcommand{\AdaptationResponsiveness}{\textit{\textcolor{DarkSeaGreen}{Responsiveness}}\xspace}
\newcommand{\AdaptationPersonalization}{\textit{\textcolor{DarkSeaGreen}{Personalization}}\xspace}
\newcommand{\AdaptationReflexivity}{\textit{\textcolor{DarkSeaGreen}{Reflexivity}}\xspace}
\newcommand{\Agency}{\textbf{\textcolor{Peach}{Agency}}\xspace}
\newcommand{\AgencyNegotiation}{\textit{\textcolor{Peach}{Negotiation}}\xspace}
\newcommand{\AgencyTransparency}{\textit{\textcolor{Peach}{Transparency}}\xspace}
\newcommand{\AgencyEmancipation}{\textit{\textcolor{Peach}{Emancipation}}\xspace}
\newcommand{\AgencyProactivity}{\textit{\textcolor{Peach}{Proactivity}}\xspace}

\begin{abstract}
Although artificial intelligence (AI) agents are increasingly proposed to support potentially longitudinal health tasks, such as symptom management, behavior change, and patient support, most current implementations fall short of facilitating user intent and fostering accountability. This contrasts with prior work on supporting longitudinal needs, both within and beyond clinical settings, where follow-up, coherent reasoning, and sustained alignment with individuals' goals are critical for both effectiveness and safety. In this paper, we draw on established clinical and personal health informatics frameworks to define what it would mean to orchestrate longitudinal health interactions with AI agents. We propose a multi-layer framework and corresponding agent architecture that operationalizes adaptation, coherence, continuity, and agency across repeated interactions. Through representative use cases, we demonstrate how longitudinal agents can maintain meaningful engagement, adapt to evolving goals, and support safe, personalized decision-making over time. Our findings underscore both the promise and the complexity of designing systems capable of supporting health trajectories beyond isolated interactions, and we offer guidance for future research and development in multi-session, user-centered health AI.
\end{abstract}
\begin{document}

\flushbottom
\maketitle
\thispagestyle{empty}

\section*{Introduction}
Large language model (LLM)–based health agents are increasingly proposed to support diverse stages of care, including triage~\cite{nedos2026artificial, gaber2025evaluating}, clinical documentation~\cite{xie2025medical, biswas2024intelligent}, decision support~\cite{maity2025large}, patient education~\cite{aydin2024large, lin2025roles}, and health coaching~\cite{jorke2025gptcoach, ong2024advancing}.
Their ability to synthesize health information and communicate directly with users has led to their exploration as a scalable means of delivering personalized guidance across a range of clinical and self-management tasks.
A number of the tasks they are designed for have the potential to evolve over time into \textit{longitudinal processes} as objectives change or remain unmet.
Chronic symptom management, for example, relies on repeated cycles of strategy testing and reflection, rather than single interactions~\cite{schulman2012processes}. 
The relevance and interpretation of health agent guidance may therefore depend on how it sustains intent and accountability over evolving circumstances, such as adapting to changing outpatient contexts and managing the flow of information during and after clinical encounters.
However, most existing health-agent systems remain \textit{reactive} and \textit{episodic}, handling each interaction as an isolated query rather than tracking evolving goals or unresolved concerns over time~\cite{peerbolte2025conversational, serugunda2025using}.
As a result, they currently offer limited support for the ongoing, longitudinal nature of health management.

Most current health agents are designed for discrete, well-defined health tasks such as patient intake~\cite{shayaninasab2024enhancing}, question answering~\cite{ayers2023comparing}, or delivering specific adaptive interventions~\cite{haag2025last, artsi2025large}.
Some digital mental health systems incorporate multi-turn or multi-session interactions, including cognitive behavioral therapy (CBT)-based chatbots~\cite{farzan2025artificial, wang2025psychological, mcfadyen2025cbt} and periodic check-in tools~\cite{wysa2026continuity}.
Yet, these approaches remain narrowly scoped.
For systems spanning multiple interactions over time, longitudinal support is frequently reduced to a conversational memory task, focused on storing, retrieving, and summarizing past exchanges~\cite{zhang2025survey}.
In doing so, sustaining user goals and accountability, which are essential for meaningful long-term outcomes, are largely neglected.
This represents a substantial gap in leveraging health agents’ capabilities for ongoing and safe longitudinal support.
Although advances in longitudinal conversational modeling and personalized health agent design continue to emerge, no generalized framework currently exists for designing longitudinal health agents that can continue work coherently across sessions when health goals remain unmet or evolve.

In this perspective, we synthesize continuity of care and personal health informatics insights to articulate a design framework for longitudinal health agents.
We introduce a multi-layer framework with various corresponding dimensions for operationalizing principles in practice.
The framework comprises four layers: (1) \coherence, which maintains consistent, contextually aware interactions through stable memory, roles, and relationships; (2) \continuity, which ensures health goals are actively stewarded through follow-up, alignment, and accountability over time; (3) \adaptation, which enables flexible responses to change via responsiveness, personalization, and reflexive reassessment; and (4) \agency, which supports the intentional negotiation of authority and responsibility through transparency, user override, emancipation, and proactive intervention.
Our framework captures continuity not only in care delivery but also in understanding and follow-up across repeated interactions, including those increasingly occurring in outpatient, preventive, and health-maintenance contexts.
As follows, we illustrate how the framework can be applied through representative use cases spanning clinical, preventive, and self-management settings.
These examples highlight how different layers become salient as individuals, clinicians, and health-system decision-making evolve over time.
Finally, we discuss key opportunities and challenges for addressing design tensions and implementation considerations inherent to longitudinal health agents, such as the maintenance of consistent longitudinal reasoning, dynamic alignment with internal and external guidelines, and adaptable agency over time, to support safe and effective real-world deployment.

\section*{Framework for persistent multi-session health agents}
\begin{figure}[!htbp]
    \centering
    \includegraphics[width=0.85\linewidth]{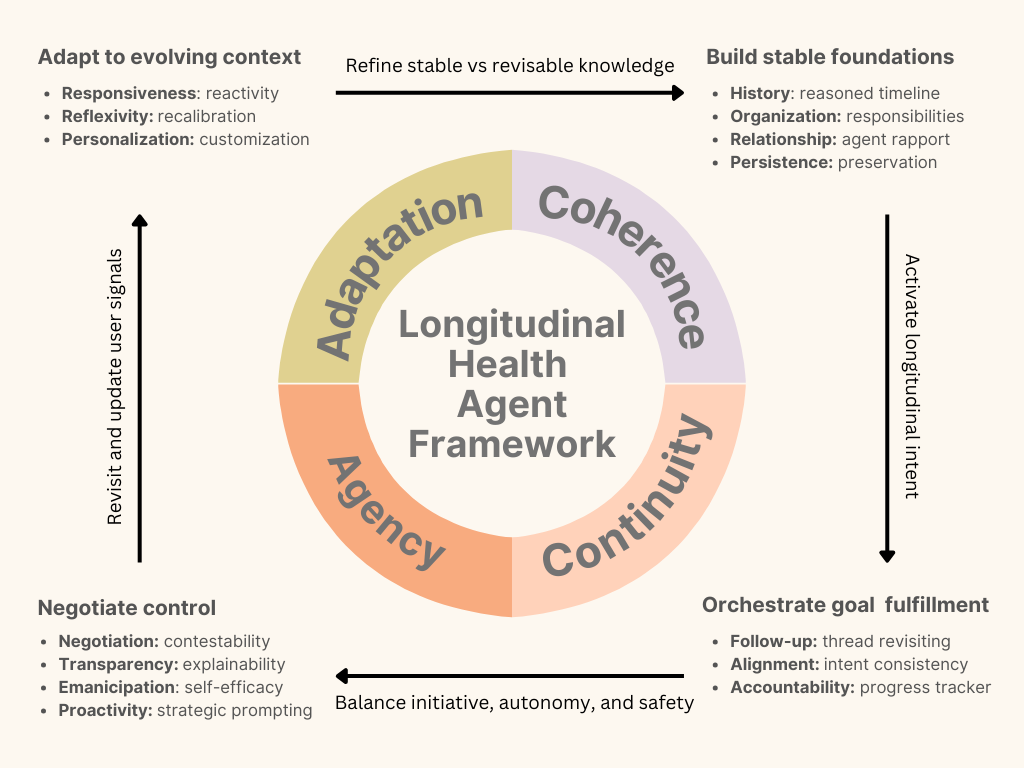}
    \caption{A four-part framework integrating \coherence, \continuity, 
    \adaptation, and \agency to support sustained health engagement over time. \Adaptation refines knowledge through \adaptationResponsiveness, \adaptationPersonalization, and \adaptationReflexivity, while \coherence builds stable interpretive foundations (\coherenceHistory, \coherenceOrganization, \coherenceRelationship, \coherencePersistence). \Continuity sustains long-term goals via \continuityFollowup, \continuityAlignment, and \continuityAccountability, and \agency calibrates control through \agencyNegotiation, \agencyTransparency, \agencyEmancipation, and \agencyProactivity. The arrows illustrate possible pathways of influence among layers (e.g., \adaptation informing \coherence by refining what remains stable versus revisable knowledge). However, in practice, these capacities are likely deeply interwoven, interacting bidirectionally and recursively across encounters rather than operating in a fixed or linear sequence.}
    \label{fig:framework}
\end{figure}
To operationalize longitudinal support in artificial intelligence (AI)-driven health agents, we synthesized insights from longitudinal clinical practice and personal health informatics, drawing on approximately 40 prior studies spanning human–computer interaction (HCI), personal health informatics, and clinical care.
This framework captures the functional and relational requirements for maintaining meaningful longitudinal engagement and supporting evolving health goals.
Our framework has four interdependent layers: \coherence, \continuity, \adaptation, and \agency (Figure~\ref{fig:framework}).
Within each layer, dimensions describe how agents can translate principles into practice.
The framework emerged iteratively through extensive discussion and reflection within our interdisciplinary team, including experts in HCI, AI, health informatics, and clinical care.
These conversations helped us refine the framework’s scope, make assumptions clear, and highlight practical design insights, ensuring it is both well-grounded and useful for supporting longitudinal health engagement.
Throughout this paper, we use the term “agent” to refer to the system providing longitudinal support, as the appropriate system architecture and its associated tradeoffs remain an open design question.
In practice, these capabilities may be implemented within a single system or distributed across multiple coordinated agents (e.g., a multi-agent system in which each layer is associated with a distinct agent).

\subsection*{\Coherence}
Classic continuity of care frameworks emphasize that care should be experienced as connected and coherent over time.
These frameworks highlight the importance of attending to evolving patient priorities, understanding relationships between triggers and symptoms, and assessing the effectiveness of past interventions as part of longitudinal support~\cite{uijen2012continuity, saultz2005interpersonal}.
Achieving this requires more than merely storing historical data; it entails reconciling longitudinal personal health management practices and evolving the longitudinal relationships among patients, clinicians, and other health stakeholders, recognizing their respective responsibilities, implementing consistent management strategies, and leveraging insights from prior encounters to guide current decisions~\cite{gray2018continuity, van2010association}.
Current health informatics systems and personalized health agents often do not fully align with these continuity-of-care principles.

Many current health tracking systems and LLM-based agents operationalize longitudinal support by storing and reusing accumulated data~\cite{peerbolte2025conversational, serugunda2025using}.
For instance, emerging LLM-based agents with multi-session capabilities often summarize or embed prior interactions, then reintroduce them at inference time to improve response relevance~\cite{zhang2022historyaware, maharana2024longterm, ge2025tremu}.
However, these systems are not designed to explicitly link individual interactions to a user’s broader goals or intentions.
Therefore, they cannot currently characterize the evolving relationships among management strategies, symptom patterns, behavioral triggers, and subsequent health outcomes.
In health contexts, where progress depends on understanding how these factors interact and change over time, this limitation reduces the system’s ability to meaningfully connect a user’s experiences, decisions, and outcomes across sessions.

When systems fail to explicitly capture user intent and data relationships, critical elements of longitudinal care are lost.
Prior assessments may not influence future decisions, evolving intervention–outcome relationships go untracked, and context-dependent priorities are overlooked.
As a result, reasoning can break down between interactions, and the system struggles to maintain a coherent and accurate narrative of an individual’s health over time.
Our framework addresses these gaps by reconceptualizing coherence as active, structured sensemaking rather than passive memory.
Historical data is not treated as a fixed input.
Instead, the \coherence layer captures interpretations, hypotheses, and reasonings that link past experiences, including both between-visit interactions and in-visit encounters, to ongoing care.

We define \coherence along four interrelated dimensions that collectively reinforce stability across interactions as seen in Table\ref{tab:coherence}.
\begin{table}[!htbp]
\centering
\small
\begin{tabularx}{\linewidth}{|p{1.7cm}|X|X|p{4.5cm}|}
\hline
\textbf{Dimension} & \textbf{Description} & \textbf{Illustrative example} & \textbf{Potential success check} \\
\hline
\CoherenceHistory\newline\cite{gray2018continuity, van2010association} 
& Store past interactions alongside agent-generated hypotheses and reasoning chains linking events to outcomes. 
& The agent links a recent symptom or behavior to previously recorded patterns and past interventions to explain likely causes.
& Whether prior interactions and reasoning are consistently and accurately incorporated into subsequent responses across sessions.\\
\hline
\CoherenceOrganization\newline\cite{uijen2012continuity} 
& Encode care entities, roles, and relationships explicitly in structured representations rather than implicit context embeddings. 
& The agent surfaces and takes into account roles (e.g., user, clinician, caregiver) and their responsibilities within an ongoing care plan.
& Whether roles, relationships, and responsibilities can be reliably identified and retrieved without contradiction or loss of structure.\\
\hline
\CoherenceRelationship\newline\cite{saultz2005interpersonal, gray2018continuity} 
& Maintain identifiable, stable interaction patterns to ensure continuity and predictable agent behavior over sessions. 
& The agent maintains a consistent interaction style and references prior discussions to reinforce continuity across sessions.
& Whether interaction patterns (e.g., tone, references, framing) remain stable and predictable across sessions.\\
\hline
\CoherencePersistence\newline\cite{ge2025tremu, wu2024longmemeval, uijen2012continuity} 
& Retain core facts, commitments, and explanatory hypotheses across updates and changing circumstances. 
& The agent retains key facts, plans, or prior conclusions across sessions and minor user context changes.
& Whether previously established information remain consistent over time, except when explicitly revised.\\
\hline
\end{tabularx}
\caption{Operationalization and evaluation of the \Coherence layer.}
\label{tab:coherence}
\end{table}

\subsection*{\Continuity}
Effective longitudinal care relies on \coherence to maintain a structured understanding of past interactions and relationships.
Complementarily, \continuity ensures that this understanding guides ongoing action, goals, and follow-up across encounters.
Clinical research defines continuity as the sustained and coordinated management of a patient’s condition across encounters~\cite{gray2018continuity, saultz2005interpersonal}.
This body of research emphasizes follow-up, responsiveness to evolving needs, prevention of fragmentation, and safety netting when symptoms persist or uncertainty remains~\cite{reynolds2018chronic, jones2019safetynetting, callen2012followup}.
Chronic care frameworks and proactive-care frameworks underscore systematic reassessment, planned follow-up, and structured management of unresolved concerns, particularly for long-term conditions such as diabetes or depression~\cite{rothman2003chronic, reynolds2018chronic, almond2009safetynetting, callen2012followup}.
These approaches highlight that meaningful progress depends on maintaining attention to goals that extend beyond a single visit while preserving safeguards against overlooked or worsening conditions.

Current digital health systems address part of this need through reminders, task lists, and just-in-time adaptive interventions (JITAIs)~\cite{li2010stagebased, nahumshani2015jitai, hsu2025jitai, bosschaerts2025jitai, lu2025digitaladherence, chen2025followupbot}.
Such systems are effective for prompting discrete behaviors (e.g., medication adherence, appointment attendance, daily logging) and improving short-term compliance~\cite{hsu2025jitai,lu2025digitaladherence}.
However, most existing systems do not represent partial progress, revisit goals without resetting them, or acknowledge forward movement when outcomes are incremental or episodic.
Continuity in these systems is often instantiated as timely prompts for predefined tasks, rather than as sustained engagement with evolving and sometimes ambiguous goals.
Many health trajectories do not follow a linear path as symptoms may fluctuate, interventions can yield partial or delayed effects, and progress may be nonlinear.
In these cases, continuity is less about completing tasks and more about maintaining orientation toward a long-term aim despite uncertainty while ensuring that unresolved issues remain visible and monitored.

Our framework reframes \continuity in agents as actively sustaining momentum across interactions.
Agents identify unresolved threads, revisit them constructively, and maintain alignment between short-term actions and broader health intentions.
Instead of treating goals as binary states (completed vs. incomplete), the agent tracks their trajectory, how they evolve, stall, or transform over time, and supports users in re-engaging without fragmentation or loss of context.

We operationalize \continuity through three interrelated dimensions as shown in Table~\ref{tab:continuity}.
\begin{table}[!htbp]
\centering
\small
\begin{tabularx}{\linewidth}{|p{1.8cm}|X|X|p{4.5cm}|}
\hline
\textbf{Dimension} & \textbf{Description} & \textbf{Illustrative example} & \textbf{Potential success check} \\
\hline
\ContinuityFollowup\newline\cite{callen2012followup, chen2025followupbot} 
& Track unresolved concerns, plans, or decisions across sessions and proactively resurface them based on prior context and risk signals rather than generic prompts. 
& The agent revisits a previously discussed concern, plan, or decision at a later time based on its relevance or status.
& Whether unresolved concerns, plans, or decisions are systematically revisited over time based on prior context without re-introduction.\\
\hline
\ContinuityAlignment\newline\cite{li2010stagebased, schulman2012processes, mamykina2015sensemaking} 
& Maintain linkage between short-term actions and evolving long-term health objectives, adjusting guidance as goals shift. 
& The agent connects immediate recommendations or behaviors to a user’s broader, evolving goals over time.
& Whether short-term recommendations remain explicitly linked to evolving long-term goals, particularly during transitions or goal changes.\\
\hline
\ContinuityAccountability\newline\cite{reynolds2018chronic, noah2018remote, hamine2015mhealth, lin2025multimodal} 
& Monitor progress over time, acknowledging partial gains, recalibration, and non-linear outcomes instead of only completed tasks. 
& The agent tracks incremental progress and setbacks and incorporates them into ongoing guidance rather than resetting progress.
& Whether progress, including partial gains and setbacks, is tracked and incorporated into subsequent guidance rather than reset or ignored.\\
\hline
\end{tabularx}
\caption{Operationalization and evaluation of the Continuity layer.}
\label{tab:continuity}
\end{table}

\subsection*{\Adaptation}
In addition to ongoing management and oversight of goals (\continuity), longitudinal support requires re-calibration of agent assumptions and behavior as health goals, circumstances, and user behaviors transition.
Prior work emphasizes that effective health tracking tools must capture and respond to changes in users' internal and external conditions~\cite{reynolds2018chronic, mamykina2015sensemaking, lin2025multimodal}.
Proactive and chronic care frameworks highlight that clinicians anticipate changes in patients’ conditions, tailor interventions accordingly, and regularly reassess goals to maintain progress and reduce risk~\cite{rothman2003chronic, reynolds2018chronic, almond2009safetynetting}.

Contemporary digital health systems handle certain aspects of this problem.
JITAIs and reinforcement learning–based health agents, for example, adjust recommendations based on contextual signals, behavioral feedback, and short-term engagement metrics to optimize intervention timing or content~\cite{nahumshani2015jitai, hsu2025jitai, haag2025last, bosschaerts2025jitai, mulani2019deep}.
However, these approaches often assume stable goals or reward structures. 
They adapt actions in response to immediate signals but rarely revisit the higher-level assumptions, inferred preferences, or causal models that guide those actions.
Systems that account for evolving user circumstances typically rely on user-initiated reflection~\cite{li2010stagebased, mamykina2015sensemaking, lin2025multimodal} and lack continuous internal representation of priorities, risk tolerance, and context-dependent assumptions.

Our framework conceptualizes \adaptation as structured recalibration across time and care contexts, extending beyond reactive adjustment.
The \adaptation layer supports not only changes in recommendations but also revision of the interpretive assumptions that guide those recommendations.
By pairing personalization with reflexivity, we introduce the notion of second-order adaptation where the agent can revisit earlier inferences about user preferences, goals, or risk tolerance and renegotiate them when emerging evidence suggests misalignment.
\Adaptation should also consider factors beyond the user to account for external changes that can significantly impact users' health journeys and interactions with the agent, such as evolving clinical guidelines, treatment standards, or regulatory updates.

We define adaptation through three interrelated dimensions as shown in Table~\ref{tab:adaptation}.

\begin{table}[!htbp]
\centering
\small
\begin{tabularx}{\linewidth}{|p{1.9cm}|X|X|p{4.5cm}|}
\hline
\textbf{Dimension} & \textbf{Description} & \textbf{Illustrative example} & \textbf{Potential success check} \\
\hline
\AdaptationResponsiveness\newline\cite{vegesna2017remote, smedslund2025remote, noah2018remote, merrill2026wearable} 
& Continuously monitor user-provided signals and surface potential external changes, adjusting reasoning and interaction logic while filtering transient noise. 
& The agent adjusts its recommendations when it detects meaningful changes in user state, behavior, or context.
& Whether meaningful changes in user state or context result in timely and appropriate updates to recommendations, without overreacting to transient variation.\\
\hline
\AdaptationPersonalization\newline\cite{abbasian2025conversational, mulani2019deep} 
& Tailor strategies and intervention policies dynamically to individual goals, behaviors, context, and longitudinal trajectories. 
& The agent tailors suggestions based on the user’s preferences, routines, and prior responses to interventions.
& Whether guidance reflects individualized patterns (e.g., preferences, routines, prior responses) and adapts as these patterns evolve over time.\\
\hline
\AdaptationReflexivity\newline\cite{mamykina2015sensemaking, mamykina2017personaldiscovery} 
& Re-evaluate assumptions, inferred preferences, and decision rules, updating internal models when longitudinal evidence shows misalignment with user states or objectives. 
& The agent revises earlier assumptions or inferred preferences when new patterns indicate they are no longer accurate.
& Whether prior assumptions or inferred preferences are revised when evidence indicates misalignment, and whether such revisions are reflected in future behavior.\\
\hline
\end{tabularx}
\caption{Operationalization and evaluation of the Adaptation layer.}
\label{tab:adaptation}
\end{table}

\noindent
\newpage
\noindent

\subsection*{\Agency}
Longitudinal care often involves shifting roles, responsibilities, and authority between patients, clinicians, and supporting systems over time.
Clinical frameworks on shared decision-making, patient activation, and collaborative care emphasize that effective care depends on explicit negotiation of responsibility~\cite{rothman2003chronic, reynolds2018chronic}.
Clinicians guide and intervene when needed, particularly when safety or clinical risk is present, while patients retain meaningful participation in decisions that affect their lives.
Informatics research similarly posits that intelligent systems should not only offer timely support but also respect users’ decision-making capacity and right to self-determination~\cite{li2010stagebased, mamykina2015sensemaking, lin2025multimodal}.

Existing health agents partially engage this tension, but often implicitly.
System outputs can be tailored to stated preferences or past behavior, and users may accept, reject, or ignore recommendations to preserve autonomy~\cite{abbasian2025conversational, su2025investigating}.
Research on recommender systems and mixed-initiative interaction often frames agency as a moment-level choice between system suggestion and user override~\cite{afroogh2024trust, sivaraman2023ignore}.
While valuable, this framing treats agency and autonomy as a transactional property of individual interactions rather than a longitudinal construct shaped cumulatively over time.

Persistent deference may overburden users during periods of vulnerability or fatigue, while persistent direction or overly affirming or “sycophantic” AI interaction patterns may gradually erode autonomy and encourage excessive dependence~\cite{afroogh2024trust, lin2025multimodal}.
Existing systems rarely model or manage these cumulative effects of guidance across sessions.
Our framework reframes agency as longitudinal calibration of authority and responsibility rather than one-off consent or momentary interaction preferences. 
The \agency layer enables the agent to intentionally adjust its degree of initiative, guidance, and intervention in response to evolving user capacity, context, and goals. 
In this way, the agent can sometimes take a very active role, such as coordinating with care teams or facilitating communication with external professionals (e.g., briefing care teams at the start of visits and integrating visit notes into memory to support coherence and continuity).
At other times, it may present itself more as a support tool, empowering the user to explore options and make decisions independently.
Over time, rather than assuming a fixed balance between system control and user choice, the agent treats agency as something negotiated and revisited across time.

We conceptualize \agency across four interrelated dimensions as shown in Table~\ref{tab:agency}.

\begin{table}[h]
\centering
\small
\begin{tabularx}{\linewidth}{|p{1.8cm}|X|X|p{4.5cm}|}
\hline
\textbf{Dimension} & \textbf{Description} & \textbf{Illustrative example} & \textbf{Potential success check} \\
\hline
\AgencyNegotiation\newline\cite{afroogh2024trust, sivaraman2023ignore} 
& Make agent initiative configurable and contestable, allowing users to question, override, or adjust guidance over time. 
& The agent allows users to modify or override its recommendations and adapts future guidance accordingly.
& Whether users can modify or override contest system recommendations, and whether these adjustments influence future interactions.\\
\hline
\AgencyTransparency\newline\cite{afroogh2024trust} 
& Expose reasoning, uncertainty, and decision logic in interpretable ways so users can calibrate trust and reliance. 
& The agent explains the reasoning behind its recommendations, including relevant inputs and uncertainty.
& Whether the system’s reasoning, inputs, and uncertainty are exposed in ways that allow users to understand and evaluate its recommendations.\\
\hline
\AgencyEmancipation\newline\cite{li2010stagebased, mamykina2015sensemaking, lin2025multimodal} 
& Support user independence by providing explanations, hints, and decision templates, gradually reducing intervention as competence increases. 
& The agent gradually reduces guidance as the user demonstrates increased confidence and independent decision-making.
& Whether user reliance on the agent decreases appropriately over time (e.g., reduced prompts) while maintaining or improving outcomes.\\
\hline
\AgencyProactivity\newline\cite{yu2025proactive, haag2025last} 
& Initiate actions or interventions appropriately while remaining aligned with user preferences and safeguards. 
& The agent initiates a check-in or suggestion when it detects potential risk or missed follow-up.
& Whether the agent initiates timely and contextually appropriate actions aligned with user goals and preferences, without excessive or misaligned intervention.\\
\hline
\end{tabularx}
\caption{Operationalization and evaluation of the Agency layer.}
\label{tab:agency}
\end{table}
\noindent

\section*{Use cases for longitudinal health agents}
To demonstrate the scope and applicability of our framework, we ground it in three representative longitudinal health contexts: (1) chronic symptom management for endometriosis, (2) post-discharge follow-up after heart failure, and (3) ongoing mental health support for anxiety.
These cases were selected to reflect distinct yet common patterns in longitudinal health, spanning chronic self-management, transitional care between clinical encounters, and sustained personal health support.
They represent variation in clinical structure, stability of goals, involvement of care teams, and levels of individual responsibility that commonly occur in real-world health trajectories.
By examining our framework across these diverse but complementary scenarios, we illustrate how \coherence, \continuity, \adaptation, and \agency can support longitudinal alignment under different forms of uncertainty, coordination demands, and evolving user needs.

\subsection*{Chronic symptom management for endometriosis}
\begin{figure}
    \centering
    \includegraphics[width=\linewidth]{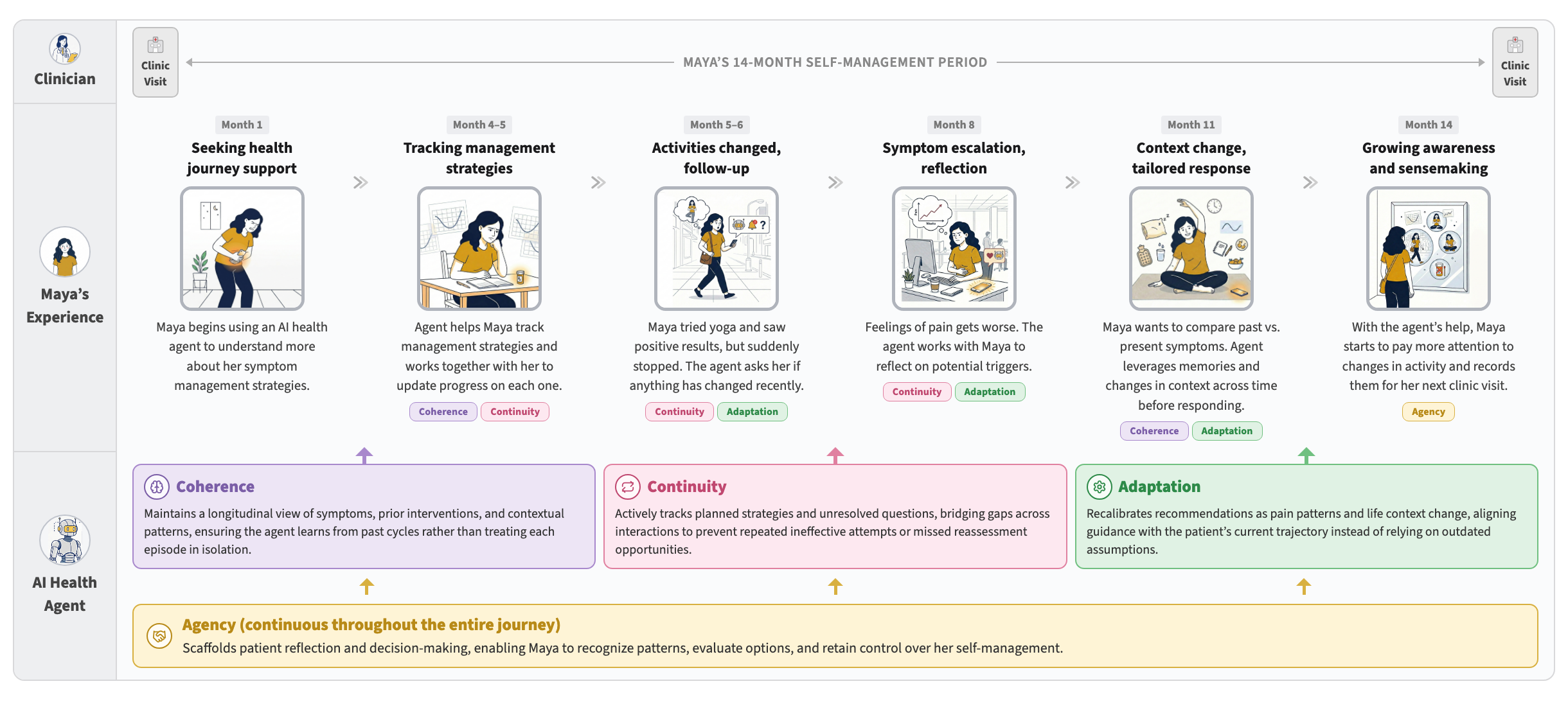}
    \caption{A potential timeline of a longitudinal health agent supporting an individual with endometriosis. The agent maintains \coherence by integrating past symptoms and interventions, ensures \continuity by actively tracking strategies and follow-ups, enables \adaptation by recalibrating guidance as symptoms and context change, and fosters \agency by scaffolding patient reflection and self-management.}
    \label{fig:usecasechronic}
\end{figure}
Endometriosis is a chronic inflammatory condition characterized by cyclic and non-cyclic pelvic pain, fatigue, gastrointestinal symptoms, and fertility concerns that often persist for years before and after diagnosis~\cite{mick2024endometriosis, becker2017treatment}.
Clinical encounters may be episodic and focused on acute escalation or procedural decisions.
As a result, patients frequently engage in extended periods of trial-and-error self-management, attempting to connect fluctuating symptoms with triggers and interventions in the absence of a clear or predictable disease progression~\cite{devan2018selfmanagement}.
Existing self-management tools, such as symptom-tracking applications, can enable users to log symptoms and relate them with cycle data~\cite{edgley2023tracking,trepanier2023apps}.
However, hypotheses about triggers, the reasoning behind trying or discontinuing a therapy, the contextual factors shaping a flare, requires users to continuously reconcile past experiences with current symptoms.
Consequently, patients may repeat ineffective strategies or delay necessary adjustments in care~\cite{becker2017treatment,devan2018selfmanagement}.

A longitudinal health agent could address some of these demands by maintaining a coherent, structured understanding of the patient’s condition over time (\coherence) as exemplified by Figure~\ref{fig:usecasechronic}.
Rather than storing only symptom counts or medication lists, the agent preserves prior hypotheses about patterns (“pain intensifies two days before menstruation when sleep is poor”), contextual explanations (“stressful work weeks amplify fatigue and pelvic discomfort”), and the reasoning behind therapeutic trials (\coherenceHistory).
When new symptoms arise, the agent situates them within this evolving model, allowing patients to refine their understanding of complex and personally variable patterns without restarting from scratch (\coherencePersistence).
Building on this structured understanding, \continuity ensures that open questions and partial trials are not lost across time.
In endometriosis management, interventions often require weeks or months to evaluate, and outcomes may be incremental rather than binary~\cite{becker2017treatment}.
The agent therefore carries forward unresolved threads, such as whether a new hormonal regimen reduced mid-cycle pain, whether dietary changes altered gastrointestinal symptoms, whether worsening fatigue warrants clinical reassessment, and reintroduces them at clinically and personally meaningful moments (\continuityFollowup).
The agent can then sustain engagement with long-term goals rather than framing progress as discrete task completion (\continuityAlignment, \continuityAccountability).
In doing so, the agent shares the burden of longitudinal oversight that patients with under-supported chronic conditions often shoulder alone.

Because disease expression and life context shift over time, \adaptation is critical.
The agent continuously monitors for meaningful changes in symptom intensity, functional impact, or life circumstances and adjusts both its recommendations and its underlying assumptions (\adaptationReflexivity).
If a previously effective strategy loses efficacy, the system revises its model rather than attributing divergence to non-adherence.
If priorities shift, for example, from pain reduction to fertility planning or vice versa, the agent recalibrates its guidance accordingly (\adaptationResponsiveness).
This form of reflexive adaptation acknowledges that both the illness and the patient’s goals evolve, and that responsible support requires periodic re-examination of earlier inferences.
Finally, sustained self-management in endometriosis requires careful attention to \agency.
Patients often become de facto experts in their own condition, yet may also experience fatigue, dismissal, or uncertainty in clinical encounters~\cite{requadt2024patient}.
A longitudinal agent can modulate its initiative over time, offering more structured guidance during periods of confusion or symptom escalation and shifting toward collaborative reflection as patients gain confidence in recognizing patterns(\agencyProactivity, \agencyEmancipation).
By making its reasoning transparent and open to contestation, the agent supports informed self-determination rather than passive compliance (\agencyTransparency).
This calibrated partnership helps preserve both autonomy and safety across the long arc of chronic illness management (\agencyNegotiation).

\subsection*{Post-discharge follow-up after heart failure}
The transition from hospital to home after heart failure hospitalization is a clinically fragile and operationally complex period~\cite{gracia2018vulnerablephase,greene2015vulnerable}.
Unlike conditions with ambiguous trajectories, heart failure management is organized around explicit physiological thresholds and defined recovery milestones, yet successful recovery depends on coordination between patients, clinicians, and caregivers~\cite{regalbuto2014discharge}.
Small lapses in adherence or monitoring can quickly lead to decompensation, making it critical that responsibilities are enacted reliably across multiple actors and settings.
Existing post-discharge tools, such as printed instructions and reminder apps, provide guidance but cannot maintain a live understanding of the patient’s recovery trajectory~\cite{heidenreich2022hfguideline}.
These tools can signal individual events, like missed medications or rising weight, but they do not interpret patterns relationally or in the context of the original discharge plan~\cite{ahrq2021readmissions,lee2016earlyfollowup,tung2017followup,lainscak2011selfcare}.
When care spans hospital, primary care, and specialty follow-up, fragmentation arises not from absent guidance, but from the absence of a persistent, integrated view that links emerging data to clinical intent.

In this context, \coherence means preserving a live model of the recovery plan and its execution.
The agent situates weight changes, medication logs, and symptom reports within the context of discharge instructions and clinical thresholds (\coherenceHistory).
Clinicians can review the same structured timeline, ensuring that the reasoning behind alerts or recommendations is clear (\agencyTransparency).
This shared representation allows the agent to highlight meaningful deviations, such as missed doses combined with rising weight, and helps the patient, agent, and clinician see patterns together (\coherenceRelationship).
By maintaining explicit links between prescribed strategies and physiological response, the agent reduces the risk that critical signals remain siloed (\coherencePersistence).
\Coherence here is less about exploratory sensemaking and more about ensuring that the biomedical logic of recovery remains intact as responsibility shifts from hospital to home (\coherenceOrganization).
Given the time-sensitive nature of post-discharge risk, \continuity can then help track whether medications are taken on schedule, follow-up appointments are completed, and symptoms are monitored, and re-engage if gaps occur (\continuityFollowup)~\cite{lee2016earlyfollowup, tung2017followup}.
\Continuity in this setting mediates between clinic and home and preserves feedback loops between the patient’s actions, physiological signals, and clinical oversight (\continuityAlignment).
In doing so, the agent ensures that care remains coordinated and coherent across settings and actors rather than fragmented into disconnected reminders (\continuityFollowup, \continuityAccountability).

As recovery progresses, \Adaptation allows the agent to adjust its support.
During the early post-discharge period, the agent may prioritize frequent check-ins and alerts to prevent clinical instability (\continuityFollowup).
Once the patient stabilizes, the agent can reduce routine check-ins but continue to flag important changes in symptoms or treatment adherence (\adaptationResponsiveness).
When clinicians modify treatment or adjust targets, the agent revises its recommendations accordingly, maintaining alignment with evolving clinical guidance (\continuityAlignment, \adaptationPersonalization).
This allows the agent to support both patients and care teams as circumstances change (\adaptationResponsiveness).
Finally, sustainable management depends on cultivating patient capacity without withdrawing necessary safeguards.
\Agency is negotiated among the patient, the agent, and the clinician (\agencyNegotiation).
In the immediate post-discharge period, the agent may assume a more directive role to reinforce adherence and safeguard against risk (\agencyProactivity).
Over time, the agent can shift toward collaborative planning by guiding the patient through reviewing symptom and adherence trends, highlighting which behaviors are contributing to recovery, and suggesting when reaching out to a clinician may be warranted based on emerging patterns (\agencyProactivity).
By making its reasoning transparent and flexible, the agent supports informed decision-making without undermining clinical authority or patient autonomy (\agencyTransparency).

\subsection*{Ongoing mental health support for anxiety and depression}
Unlike highly structured medical conditions, mental health trajectories often involve a mix of formal clinical encounters, intermittent therapy, medication adjustments, and extended periods of personal health management~\cite{balaskas2021emi}.
Clinical involvement may be sparse, irregular, or entirely absent, leaving patients to interpret subtle changes in sleep, motivation, social engagement, or mood on their own~\cite{torous2021digital}.
At the same time, when clinicians or therapists are involved, care is structured around specific goals or interventions that must be coordinated with daily coping strategies.
Existing digital mental health tools, including both clinically integrated systems and consumer self-help applications, however, are typically designed around isolated interactions.
Mood-tracking apps, conversational agents, and crisis services capture snapshots of emotional state or intervene in acute moments without visibility into the trajectory that preceded escalation~\cite{torous2021digital}.
This intermediate space that can be structured and self-directed creates unique challenges for longitudinal support, where goal stability, individual responsibility, and external stakeholder involvement vary over time.

Through the \coherence layer, the agent could construct a cumulative and evolving understanding of an individual’s emotional patterns, triggers, coping responses, and therapeutic goals.
Anxiety and depression are context-sensitive and probabilistic with the same behavior or mood shift signaling different states depending on circumstances~\cite{haaker2015deficient}.
\Coherence would allow the agent to differentiate these patterns by connecting current reports to prior episodes, contextual stressors, and previously attempted strategies (\coherenceHistory).
\Coherence could also encompass memory of user preferences for engagement, such as how the user responds to prompts, what forms of validation are helpful, and when encouragement or gentle guidance is appropriate (\coherenceRelationship).
If clinicians are involved, the agent can represent their guidance alongside self-management strategies, ensuring that both patient and care team share a coherent understanding of progress and priorities (\coherenceOrganization, \coherencePersistence).
The \continuity layer could then help maintain presence across the periods between therapy or psychiatric appointments (\continuityFollowup).
The agent can monitor patterns that historically precede relapse, such as changes in communication, disrupted routines, or missed self-care behaviors, and intervene proactively (\agencyProactivity).
Rather than issuing generic prompts, it could situate current difficulties within prior cycles of recovery and relapse, reinforcing that temporary setbacks fit within a broader trajectory (\continuityAlignment). 
\Continuity in this domain serves both preventive and stabilizing functions by sustaining engagement during stable periods and enabling earlier recognition of destabilization during vulnerable ones (\continuityAccountability, \adaptationResponsiveness).

Users' emotional states can shift rapidly, and interventions that are helpful in one phase may feel burdensome or misaligned in another.
The \adaptation layer could recalibrate tone, structure, and intensity of support to respond to these shifts in real time (\adaptationResponsiveness).
During a period of stability, the agent might emphasize goal-setting, skill development, or reflective exercises (\adaptationPersonalization).
During a period of distress, it could simplify interactions, prioritize grounding techniques, or encourage timely contact with clinicians (\adaptationResponsiveness).
Because mental health symptoms themselves can impair motivation and executive function, \Adaptation involves dynamically adjusting expectations in real time with symptom severity, cognitive capacity, and contextual stressors (\continuityAlignment).
With potential reductions in decision-making capacity, making some forms of choice or responsibility can be overwhelming.
However, complete removal of agency can foster helplessness~\cite{hindmarch2013depression}.
The \Agency layer in a longitudinal health agent negotiates this balance, dynamically calibrating the locus of control (\agencyNegotiation).
This negotiation can be informed by longitudinal context where the agent can track patterns of decision-making capacity over time, observe when the individual reliably responds to prompts, initiates coping behaviors independently, or benefits from structured scaffolding (\adaptationReflexivity).
The agent could strategically adjust who drives action and provide support that aligns with both immediate capability and long-term self-management goals (\agencyProactivity, \agencyEmancipation).

\section*{Implications, challenges, and future directions for longitudinal health agents}
This section outlines key tensions, challenges, and open questions that arise when designing and evaluating longitudinal health agents.

\subsection*{Design tensions in longitudinal agent systems}
Below we highlight core design tensions that emerge when balancing competing longitudinal properties within agent systems.

\textbf{\textit{Preserving meaning while allowing change}}
Our framework reveals a tension between information stability and necessary revision through its \coherence layer and \adaptation layers.
Overemphasizing coherence risks solidifying tentative assumptions into misleading long-term beliefs, whereas prioritizing adaptation too heavily can fragment the user’s narrative.
This tension could be worsened by the risk that initially reasonable outputs may become problematic if treated as persistent authority without reassessment, especially as external knowledge (e.g., clinical guidelines and regulation) evolves.
Agents must therefore track not just relevant information but the status and basis of interpretations in order to distinguish durable knowledge from revisable hypotheses and incorporating oversight or verification when needed.
These challenges can complicate evaluation.
While storage is easy to measure, it is far harder to assess whether systems retain and revise the right information over time, particularly as failures may stem from outdated knowledge, flawed reasoning, or appropriate uncertainty.
Ultimately, ensuring alignment requires longitudinal evaluation of coherence drift and user perceptions of whether revisions are justified, transparent, and responsive to changing evidence.

\textbf{\textit{Sustaining direction and momentum}}
A parallel tension exists between short-term task execution and long-term trajectory.
Discrete tasks, such as trying a new intervention, monitoring symptoms, or scheduling follow-up care, remain essential, yet their significance is often realized only through accumulation, comparison, and reflection across extended periods.
Current evaluations of health agents tend to prioritize diagnostic accuracy or task completion within single interactions~\cite{ong2024advancing, npjDigMed2025bias}, but this narrow focus can misclassify pauses or regressions as failure, despite nonlinear progress being typical in health contexts~\cite{schulman2012processes}.
Supporting \continuity therefore requires representing goals as evolving pathways rather than binary endpoints. 
Agents must help users see how incremental effort, reflection, and recalibration contribute to a broader direction, even when outcomes remain uncertain.
Evaluating this capability cannot rely solely on short-term adherence metrics alone, as system might optimize isolated task completion without sustaining engagement with long-term goals.
Demonstrating true continuity may require following users across extended periods to examine whether unresolved concerns are revisited, whether partial gains are acknowledged and built upon, and whether users maintain orientation toward meaningful objectives despite interruptions.

\textbf{\textit{Longitudinal proactivity without erosion of autonomy}}
Treating \agency as a longitudinal variable introduces a tension between proactive support and respect for autonomy, as independence is not fixed but shifts with users’ conditions, circumstances, and goals.
Rather than a linear progression toward full independence, agency becomes an ongoing negotiation of roles and authority, requiring systems to make their level of initiative both legible and adjustable.
Agents must balance when to intervene and when to step back, since persistent over-direction can undermine confidence while excessive deference may leave risks unaddressed.
Of note, autonomy may also include the user’s ability to retract or delete previously shared data, which can reshape the agent’s contextual understanding and limit its capacity to provide longitudinally informed support.
This underscores the need for longitudinal systems to handle such missing data transparently, recalibrate appropriately, and communicate any resulting uncertainty or limitations in guidance.
These cumulative effects are difficult to capture in traditional evaluation paradigms.
Although current personalization methods often rely on moment-level behavioral data like adherence or engagement~\cite{mulani2019deep}, such metrics do not reveal whether users become more capable, dependent, or disengaged over time.
Assessing longitudinal agency instead requires combining behavioral data with repeated measures of perceived competence, trust, and autonomy, as well as examining whether proactivity improves safety without diminishing user control, including how other stakeholders perceive the agent’s involvement in care.

\subsection*{Future research directions and implementation considerations}
Emerging agent frameworks already include useful building blocks for longitudinal health agents such as persistent memory, evolving system state, and proactive behaviors like periodic triggers and background processes~\cite{yu2025proactive, maharana2024longterm, ge2025tremu}.
However, these capabilities alone are not sufficient to develop a longitudinal health agent as they do not track changing health goals, separate tentative ideas from stable knowledge, and calibrate their control around user circumstances overtime.
Addressing this will likely require more structured representations of longitudinal state, such as maintaining time-indexed goal hierarchies, explicitly labeled hypotheses (e.g., tentative vs. confirmed), and relational links between symptoms, interventions, and outcomes.
In parallel, systems will need mechanisms to ground outputs in verified medical knowledge and to maintain calibrated estimates of uncertainty and clear provenance trails as these internal representations evolve over time.

In practice, these systems will also likely not operate in isolation.
Longitudinal agents must coordinate across components and stakeholders, whether implemented as multi-agent systems or integrated into existing healthcare infrastructure.
This includes interfacing with electronic health records (EHRs) to incorporate validated clinical data and communicate longitudinal insights back to care teams.
Such integration raises challenges in maintaining shared representations, avoiding fragmentation, and ensuring that reasoning remains coherent across systems, encounters, and stakeholders.
Differences between patient-reported data and clinical records, delays in updates, and mismatched data formats can all introduce inconsistencies that grow over time if they are not actively managed.

On the other hand, the system might not always be able to rely on clinician-mediated validation or communication as users might engage in independent self-management practices irrespective of clinical access.
In these settings, longitudinal health agents might be more focused on supporting individuals in forming, revising, and maintaining their own understanding of their health trajectories over time.
This shifts key design requirements beyond coordination with external stakeholders toward a more demanding need for internal mechanisms of self-calibration, including robust communication of uncertainty, safeguards against over-reliance on system outputs, and support for situating short-term experiences within longer-term patterns.

The accumulation of sensitive health data over time raises important privacy and governance concerns.
Longitudinal agents must support ongoing consent, clear data ownership, and alignment with regulatory frameworks (e.g., HIPAA, GDPR). 
This may necessitate features for selective memory revision or deletion.
As agents adopt more proactive roles, however, their actions may end up depending on incomplete or outdated data, introducing substantial safety, ethical, and clinical risks.
Thus, ensuring safe deployment will likely require oversight, auditing, and governance mechanisms such as human-in-the-loop oversight, auditable reasoning, and clear boundaries between assistive support and clinical decision-making.
While the allocation of responsibility and liability will depend on evolving regulatory and legal frameworks across jurisdictions, a near-term design priority is for longitudinal agents acting on behalf of users to be explicit about when, why, and under whose authority they act, and to retain sufficient context to enable subsequent review and audit.
We see this as a shared challenge for designers, developers, clinicians, and policymakers to navigate collectively as these systems mature.

In Table \ref{tab:implementation} below, we outline open questions for future research.

\begin{table}[h]
\centering
\small
\renewcommand{\arraystretch}{1.2}
\begin{tabularx}{\linewidth}{|p{4cm}|X|}
\hline
\textbf{Longitudinal Health Agent Design} & \textbf{Open questions}\\
\hline
Persistent memory and structured context (\coherence) &
\begin{minipage}[t]{\hsize}
\raggedright
\begin{itemize}[leftmargin=*, noitemsep, topsep=0pt]
    \item How can agents represent and update structured longitudinal knowledge to maintain coherence across sessions while allowing safe revision?
    \item What are effective ways to encode relationships, roles, and hypotheses for evolving health contexts?
\end{itemize}
\end{minipage} \\
\hline
Tracking evolving goals (\continuity) &
\begin{minipage}[t]{\hsize}
\raggedright
\begin{itemize}[leftmargin=*, noitemsep, topsep=0pt]
    \item How can agents dynamically capture partial progress and non-linear trajectories?
    \item What methods best link short-term tasks to long-term objectives and maintain alignment with shifting user priorities?
\end{itemize}
\end{minipage} \\
\hline
Reflexive adaptation (\adaptation) &
\begin{minipage}[t]{\hsize}
\raggedright
\begin{itemize}[leftmargin=*, noitemsep, topsep=0pt]
    \item How can agents detect misalignment between past inferences and current internal and external evidence, and update strategies without disrupting continuity?
    \item How should second-order adaptation be evaluated longitudinally?
\end{itemize}
\end{minipage} \\
\hline
Longitudinal agency calibration (\agency) &
\begin{minipage}[t]{\hsize}
\raggedright
\begin{itemize}[leftmargin=*, noitemsep, topsep=0pt]
    \item How can agents negotiate initiative over time, balancing proactive support and user autonomy?
    \item What metrics can capture user empowerment, trust, and decision-making confidence across extended use?
\end{itemize}
\vspace{1pt}
\end{minipage} \\
\hline
Evaluation of longitudinal outcomes &
\begin{minipage}[t]{\hsize}
\raggedright
\begin{itemize}[leftmargin=*, noitemsep, topsep=0pt]
    \item What study designs, metrics, and qualitative approaches can capture performance across coherence, continuity, adaptation, and agency over time?
    \item How can longitudinal impact on user trajectories be assessed systematically?
\end{itemize}
\end{minipage} \\
\hline
\end{tabularx}
\caption{Potential research agenda for longitudinal health agents. The table highlights open questions for each key capability of the framework.}
\label{tab:implementation}
\end{table}

\section*{Discussion}
LLM-based health agents are increasingly proposed as tools to support complex and evolving care processes.
However, most current systems remain episodic, reactive, and limited to discrete tasks.
In this perspective, we argue that meaningful health support requires more than accurate responses within isolated encounters.
It depends on structured, longitudinal stewardship of goals, interpretations, and responsibilities over time.
Drawing on insights from continuity of care and personal health informatics, we introduce a four-layer framework that conceptualizes longitudinal support as an active design challenge rather than a byproduct of memory or personalization.
This framework shifts attention away from static data retention and short-term adherence toward sustained intent, evolving interpretation, negotiated authority, and accountable follow-up across interactions.
The use cases demonstrate both the potential and the complexity of this approach.
Longitudinal agents may enhance coherence, engagement, and personalization, but they also raise new concerns related to safety, calibration, and the appropriate delegation of responsibility.
Addressing these challenges will require new evaluation paradigms, system architectures, and interface designs that explicitly support ongoing, multi-session engagement, rather than treating continuity as an extension of single encounters.

\section*{Data availability}
No datasets were generated or analyzed during the current study.

\section*{Code availability}
No code was generated or analyzed during the current study.

\bibliography{sample}

\section*{Acknowledgements}
Research reported in this publication was supported by the National Library of Medicine Training Grant and the Columbia University Research Stabilization Grant.

\section*{Author information}
\subsection*{Contributions}
G.L., N.E., and X.X. conceived the core ideas described in this study. G.L. and R.J. wrote the initial draft of the manuscript. All authors contributed to the ideas described herein, reviewed and approved the final version of the manuscript.

\section*{Ethics declarations}
\subsection*{Competing interests}
The authors declare no competing interests.

\end{document}